\documentclass[10pt,twocolumn,letterpaper]{article}
\pdfoutput=1
\usepackage{cvpr}
\usepackage{times}
\usepackage{epsfig}
\usepackage{graphicx}
\usepackage{amsmath}
\usepackage{amssymb}

\usepackage{bbm}

\usepackage[breaklinks=true,bookmarks=false]{hyperref}

\cvprfinalcopy 

\def\cvprPaperID{****} 

\begin{document}

\title{Video Summarization by Learning from Unpaired Data}

\author{Mrigank Rochan and Yang Wang\\
University of Manitoba\\
{\tt\small \{mrochan, ywang\}@cs.umanitoba.ca}
}

\maketitle

\begin{abstract}
	We consider the problem of video summarization. Given an input raw video, the goal is to select a small subset of key frames from the input video to create a shorter summary video that best describes the content of the original video. Most of the current state-of-the-art video summarization approaches use supervised learning and require labeled training data. Each training instance consists of a raw input video and its ground truth summary video curated by human annotators. However, it is very expensive and difficult to create such labeled training examples. To address this limitation, we propose a novel formulation to learn video summarization from unpaired data. We present an approach that learns to generate optimal video summaries using a set of raw videos ($V$) and a set of summary videos ($S$), where there exists no correspondence between $V$ and $S$. We argue that this type of data is much easier to collect. Our model aims to learn a mapping function $F : V \rightarrow S$ such that the distribution of resultant summary videos from $F(V)$ is similar to the distribution of $S$ with the help of an adversarial objective. In addition, we enforce a diversity constraint on $F(V)$ to ensure that the generated video summaries are visually diverse. Experimental results on two benchmark datasets indicate that our proposed approach significantly outperforms other alternative methods.


\end{abstract}

\section{Introduction}\label{sec:intro}

In recent years, there has been a phenomenal surge in videos uploaded online everyday. With this remarkable growth, it is becoming difficult for users to watch or browse these online videos efficiently. In order to make this enormous amount of video data easily browsable and accessible, we need automatic video summarization tools. The goal of video summarization is to produce a short summary video that conveys the important and relevant content of a given longer video. Video summarization can be an indispensable tool that has potential applications in a wide range of domains such as video database management, consumer video analysis and surveillance \cite{xiong2006unified}.

Video summarization is often formulated as a subset selection problem. In general, there are two types of subset selection in video summarization: (i) key frames selection, where the goal is to identify a set of isolated frames \cite{gong14_nips,lee2012discovering,liu2010hierarchical,mahasseni17_cvpr,mundur2006keyframe,ours18_eccv,zhang16_eccv}; and (ii) key shots selection, where the aim is to identify a set of temporally continuous interval-based segments or subshots \cite{laganiere2008video,lu2013story,nam2002event,ngo2003automatic_iccv}. In this paper, we treat video summarization as a key frame selection problem. A good summary video should contain video frames that satisfy certain properties. For example, the selected frames should capture the key content of the video \cite{gong14_nips,khosla2013large,ngo2003automatic_iccv}. In addition, the selected frames should be visually diverse \cite{gong14_nips,mahasseni17_cvpr,zhang16_eccv}. 

Both supervised and unsupervised learning approaches have been proposed for video summarization. Most unsupervised methods \cite{khosla2013large,kim2014reconstructing_cvpr,lee2012discovering,lu2013story,mahasseni17_cvpr,ngo2003automatic_iccv,panda2017collaborative,Song_2015_CVPR,zhou2018_aaai,ours18_eccv, zhang2018_eccv} use hand-crafted heuristics to select frames in a video. The limitation of such approaches is that it is difficult to come up with the heuristics that are sufficient for generating good summary videos. In contrast, supervised methods \cite{gong14_nips,gygli14_eccv,gygli15_cvpr,zhang16_cvpr,zhang16_eccv,zhao2017_acmm} learn from training data with user-generated summary videos. Each instance of the training data consists of a pair of videos -- a raw input video and its corresponding ground truth summary video created by humans. From such training data, these supervised methods learn to map from a raw input video to a summary video by mimicking how humans create summary videos. Supervised methods can implicitly capture cues used by humans that are difficult to model via hand-crafted heuristics, so they tend to outperform unsupervised methods.

\begin{figure*}[t]
	\center
	\includegraphics[width=0.9\textwidth]{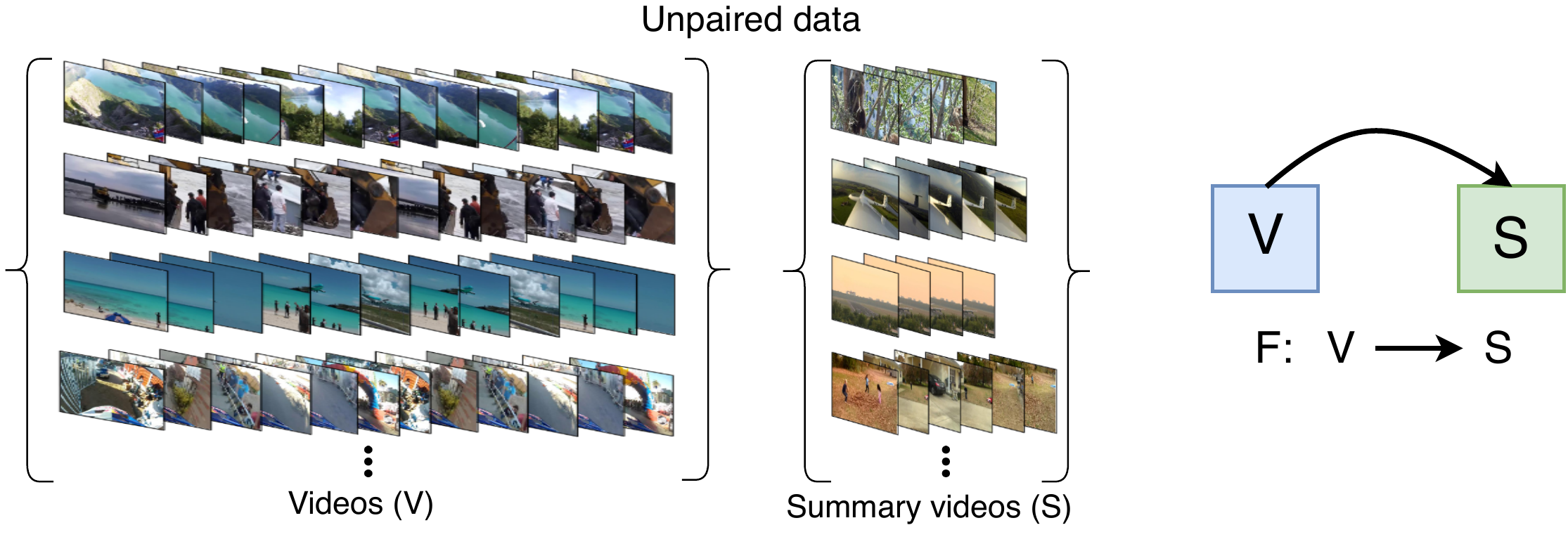}
	\caption{ Learning video summarization from unpaired data. Given a set of raw videos \{$v_i$\}$_{i=1}^M$ ($v \in V$) and real summary videos \{$s_j$\}$_{j=1}^N$ ($s \in S$) such that there exists no matching/correspondence between the instances in $V$ and $S$, our aim is to learn a mapping function $F : V \rightarrow S $ (right) linking two different domains $V$ and $S$. The data are \textit{unpaired} because the summary set $S$ does not include ground truth summary videos for raw videos in $V$, and vice versa.}
	\label{fig:problem}
\end{figure*}

A major limitation of supervised video summarization is that it relies on labeled training data. Common datasets in the community are usually collected by asking human annotators to watch the input video and select the key frames or key shots. This annotation process is very expensive and time-consuming. As a result, we only have very few benchmark datasets available for video summarization in the computer vision literature. Moreover, each dataset usually only contains a small number of annotated data (see Table \ref{table:datasets}).

To address the flaws of supervised learning, we propose\textit{ a new formulation of learning video summarization from unpaired data}. Our key insight is that it is much easier to collect unpaired video sets. First of all, raw videos are easily accessible as they are abundantly available on the Internet. At the same time, good summary videos are also readily available in large quantities. For example, there are lots of sports highlights, movie trailers, and other professionally edited summary videos available online. These videos can be treated as ground truth summary videos. The challenge is that these professionally curated summary videos usually do not come with their corresponding raw input videos. In this paper, we propose to solve video summarization by learning from such unpaired data~(Fig.~\ref{fig:problem} (left)). We assume that our training data consist of two sets of videos: one set of raw videos ($V$) and another set of human created summary videos ($S$). However, there exists no correspondence between the videos in these two sets, i.e. the training data are \textit{unpaired}. In other words, for a raw video in $V$, we may not have its corresponding ground truth summary video in $S$, and vice versa. 

We propose a novel approach to learn video summarization from unpaired training data. Our method learns a mapping function (called the \emph{key frame selector}) $F : V \rightarrow S $ (Fig.~\ref{fig:problem} (right)) to map a raw video $v\in V$ to a summary video $F(v)$. It also trains a \emph{summary discriminator} that tries to differentiate between a generated summary video $F(v)$ and a real summary video $s \in S$. Using an adversarial loss~\cite{goodfellow14_nips}, we learn to make the distribution of generated summary videos $F(v)$ to be indistinguishable from the distribution of real summary videos in $S$. As a result, the mapping function $F$ will learn to generate a realistic summary video for a given input video. We also add more structure to our learning by introducing a reconstruction loss and a diversity loss on the output summary video $F(v)$. By combining these two losses with the adversarial loss, our method learns to generate meaningful and visually diverse video summaries from unpaired data.

In summary, our contributions of this paper include: (i) a new problem formulation of learning video summarization from unpaired data, which consists of a set of raw videos and a set of video summaries that do not share any correspondence; (ii) a deep learning model for  video summarization that learns from unpaired data via an adversarial process; (iii) an extensive empirical study on benchmark datasets to demonstrate the effectiveness of the proposed approach; and (iv) an extension of our method that introduces the partial supervision to improve the summarization performance.


\section{Related Work}\label{sec:related}
With the explosive increase in the amount of online video data, there has been a growing interest in the computer vision community on developing automatic video summarization techniques. Most of the prior approaches fall in the realm of unsupervised and supervised learning.

Unsupervised methods \cite{chu2015video,de2011vsumm,kang2006space,kim2014reconstructing_cvpr,lee2012discovering,liu2002optimization,lu2013story,ma2002user,ngo2003automatic_iccv,panda2017collaborative,potapov14_eccv,Song_2015_CVPR,zhang1997integrated,zhao2014quasi} typically use hand-crafted heuristics to satisfy certain properties (e.g. diversity, representativeness) in order to create the summary videos. Some summarization methods also provide weak supervision through additional cues such as web images/videos \cite{cai2018weakly,khosla2013large,kim2014reconstructing_cvpr,Song_2015_CVPR} and video category information \cite{panda2017weakly,potapov14_eccv} to improve the performance.

Supervised methods \cite{gong14_nips,gygli14_eccv,gygli15_cvpr,Li_2018_ECCV,mahasseni17_cvpr,ours18_eccv,Sharghi_2018_ECCV,zhang16_cvpr,zhang16_eccv,zhang2018_eccv,zhang2018_bmvc,zhang2018_arxiv,zhou2018_aaai} learn video summarization from labeled data consisting of raw videos and their corresponding ground-truth summary videos. Supervised methods tend to outperform unsupervised ones, since they can learn useful cues from ground truth summaries that are hard to capture with hand-crafted heuristics. Although supervised methods are promising, they are limited by the fact that they require expensive labeled training data in the form of videos and their summaries (i.e., paired data). In this paper, we propose a new formulation for video summarization where the algorithm only needs unpaired videos and summaries (see Fig. \ref{fig:problem}(left)) for training. The main advantage of such unpaired data is that it is much easier to collect them.

Recent methods treat video summarization as a structured prediction task \cite{mahasseni17_cvpr,ours18_eccv,zhang16_eccv,zhang2018_eccv,zhou2018_aaai,zhao2017_acmm,zhao2018_cvpr,zhou2018_bmvc}. In particular, our formulation aligns with Rochan \etal \cite{ours18_eccv} that models the video summarization as a sequence labeling problem. Unlike contemporary methods  \cite{zhang16_eccv,mahasseni17_cvpr,zhao2017_acmm,zhao2018_cvpr,zhang2018_eccv} that use recurrent models, Rochan \etal \cite{ours18_eccv} propose fully convolutional sequence model which is efficient and allows better GPU parallelization. However, the major limitation of their method is that it is fully supervised and relies on paired training data. In contrast, we aim to learn video summarization using videos and summaries that have no matching information (i.e., unpaired data).

Lastly, our notion of learning from unpaired data is partly related to recent research in image-to-image translation \cite{almahairi2018_icml,chen2018_cvpr,yi2017_iccv,CycleGAN2017}. These methods learn to translate an input image from one domain to an output image in another domain without any paired images from both domains during training. However, there are major technical differences between these methods and ours. They typically employ two-way generative adversarial networks (GANs) with cycle consistency losses, whereas our method is an instance of standard GANs \cite{goodfellow14_nips} with losses designed to solve video summarization. Moreover, their formulation is limited to unpaired learning in images. To the best of our knowledge, this paper is the first work on unpaired learning in video analysis, in particular video summarization.

%

\section{Our Approach}\label{sec:approach}

\subsection{Formulation}\label{sec:formulation}
We are given an unpaired dataset consisting of a set of raw videos \{$v_i$\}$_{i=1}^M$ and a set of real summary videos \{$s_j$\}$_{j=1}^N$, where $v_i \in V$ and $s_j \in S$. We define the data distribution for $v$ and $s$ as $v \sim p_{data}(v)$ and $s \sim p_{data}(s)$, respectively. Our model consists of two sub-networks called the \emph{key frame selector network} ($S_{K}$) and the \emph{summary discriminator network} ($S_{D}$). The key frame selector network is a mapping function $S_K : V \rightarrow S $ between the two domains $V$ and $S$ (see Fig. \ref{fig:problem}). Given an input video $v\in V$, the key frame selector network ($S_{K}$) aims to select a small subset of $k$ key frames of this video to form a summary video $S_{K}(v)$. The goal of the summary discriminator network ($S_D$) is to differentiate between a real summary video $s\in S$ and the summary video $S_{K}(v)$ produced by the key frame selector network $S_K$. Our objective function includes an adversarial loss, a reconstruction loss and a diversity loss. We learn the two networks $S_{K}$ and $S_{D}$ in an adversarial fashion. In the end, $S_{K}$ learns to output an optimal summary video for a given input video. In practice, we precompute the image feature of each frame in a video. With a little abuse of terminology, we use the term ``video'' to also denote the sequence of frame-level feature vectors when there is no ambiguity based on the context.

\subsection{Network Architecture}

\begin{figure*}[ht]
	\center
	\includegraphics[width=0.9\textwidth]{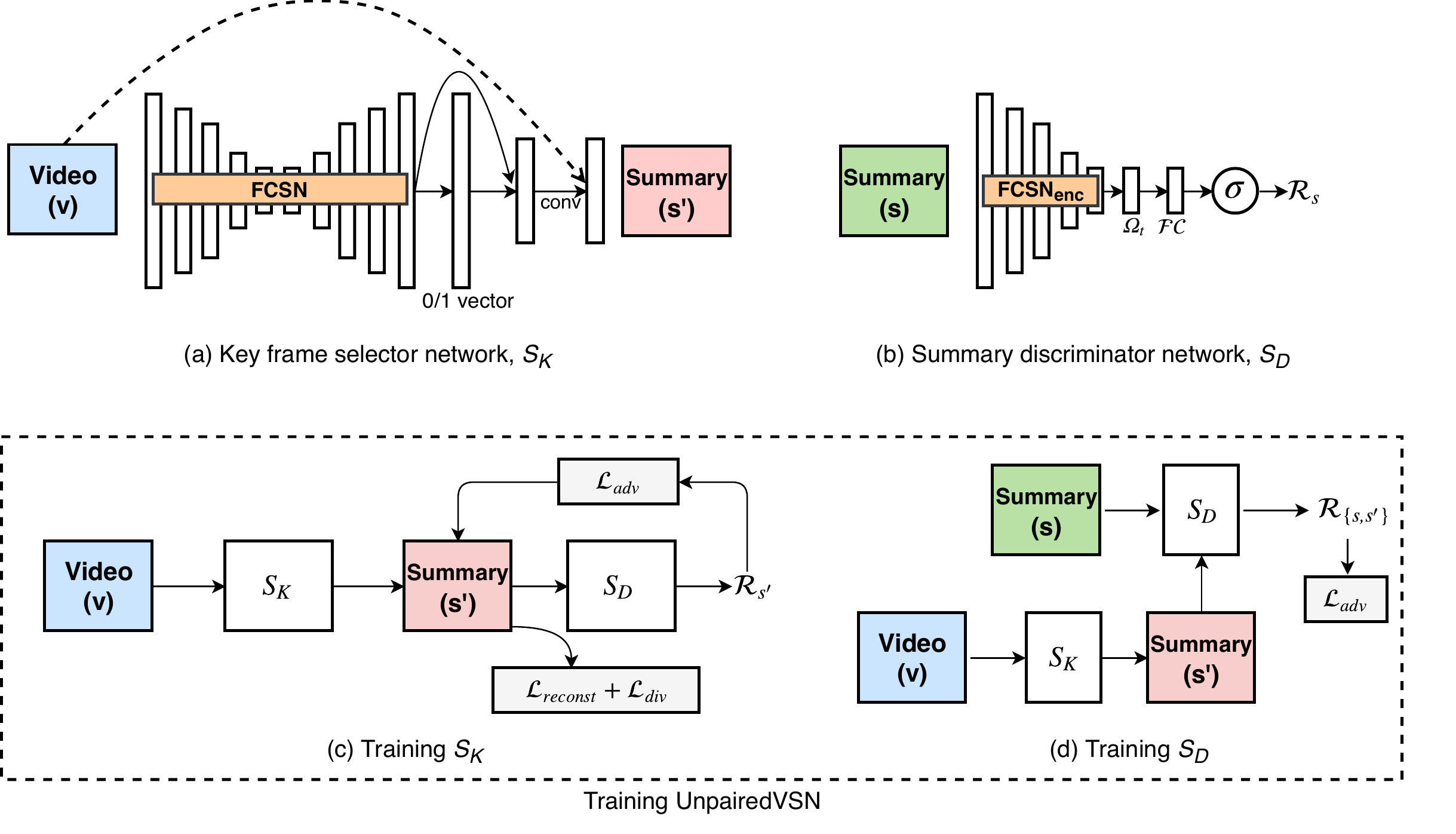}
	\caption{Overview of our proposed model. (a) Network architecture of the key frame selector network $S_K$. It takes a video $v$ and produces its summary video $s^{\prime}$ (i.e., $S_K(v)$) by selecting $k$ key frames from $v$. The backbone of $S_K$ is FCSN~\cite{ours18_eccv}. We also introduce a skip connection from the input to retrieve the frame-level features of $k$ key frames selected by $S_K$. (b) Network architecture of the summary discriminator network $S_D$. It differentiates between an output summary video $s^{\prime}$ and a real summary video $s$. $S_D$ consists of the encoder of FCSN ($FCSN_{enc}$), followed by a temporal average pooling ($\Omega_t$) and sigmoid ($\sigma$) operations. In (c) and (d), we show the training scheme of $S_K$ and $S_D$, respectively. $S_K$ tries to produce video summaries that are indistinguishable from real video summaries created by humans, whereas $S_D$ tries to differentiate real summary videos from the summaries produced by $S_K$. As mentioned in Sec.~\ref{sec:formulation}, there is no correspondence information available to match raw videos and summary videos in the training data.}
	\label{fig:network}
\end{figure*}

The \textit{key frame selector network} ($S_K$) in our model takes a video with $T$ frames as the input and produces the corresponding summary video with $k$ key frames. We use the fully convolutional sequence network (FCSN) \cite{ours18_eccv}, an encoder-decoder fully convolutional network, to select key frames from the input video. FCSN encodes the temporal information among the video frames by performing convolution and pooling operations in the temporal dimension. This enables FCSN to extract representations that capture the inter-frame structures. The decoder of FCSN consists of several temporal deconvolution operations which produces a vector of prediction scores with the same length as the input video. Each score indicates the likelihood of the corresponding frame being a key frame or non-key frame. Based on these scores, we select $k$ key frames to form the predicted summary video. In order to define the reconstruction loss used in the learning (see Sec.~\ref{sec:learning}), we apply convolution operations on the decoded feature vectors of these $k$ key frames to reconstruct the corresponding feature vectors in the input video. We also introduce a skip connection that retrieves the frame-level feature representation of the selected $k$ key frames, which we merge with the reconstructed features of the $k$ key frames. Fig.~\ref{fig:network}~(a) shows the architecture of $S_K$.

The \textit{summary discriminator network} ($S_D$) in our model takes two kinds of input: (1) the summary videos produced by $S_K$ for the raw videos in $V$; and (2) the real summary videos in $S$. The goal of $S_D$ is to distinguish between the summaries produced by $S_K$ and the real summaries. We use the encoder of FCSN \cite{ours18_eccv} to encode the temporal information within the input summary video. Next, we perform a temporal average pooling operation ($\Omega_t$) on the encoded feature vectors to obtain a video-level feature representation. Finally, we append a fully connected layer ($\mathcal{FC}$), followed by a sigmoid operation ($\sigma$) to obtain a score ($\mathcal{R}_s$) indicating whether the input summary video is a real summary or a summary produced by $S_K$. Let $s$ be an input summary video to $S_D$, we can express the operations in $S_D$ by Eq.~\ref{eq:sd}. The network architecture of $S_D$ is shown in Fig.~\ref{fig:network}~(b).
\begin{equation}\label{eq:sd}
\mathcal{R}_s = S_D(s) = \sigma\,(\mathcal{FC}\,(\Omega_t\,(FCSN_{enc}(s))))
\end{equation}

\subsection{Learning}\label{sec:learning}
Our learning objective includes an adversarial loss \cite{goodfellow14_nips}, a reconstruction loss, and a diversity loss.

\noindent{\textbf{Adversarial Loss}}: This loss aims to match the distribution of summary videos produced by the key frame selector network $S_K$ with the data distribution of real summary videos. We use the adversarial loss commonly used in generative adversarial networks~\cite{goodfellow14_nips}:
\begin{eqnarray}\label{eq:l_adv}
\mathcal{L}_{adv}(S_D,S_K)&=&\mathbb{E}_{s \sim p_{data}(s)} [\log S_D(s)]\\
&&+\mathbb{E}_{v \sim p_{data}(v)}[\log (1-S_D(S_K(v)))]\nonumber
\end{eqnarray}
where $S_K$ aims to produce summary videos $S_K(v)$ that are close to real summary videos in domain $S$, and $S_D$ tries to differentiate between output summary videos $\{S_K(v): v \in V\}$ and real summary videos $\{s: s\in S\}$. A minimax game occurs between $S_K$ and $S_D$, where $S_K$ pushes to minimize the objective and $S_D$ aims to maximize it. This is equivalent to the following: 
\begin{equation}
\min_{S_K}\max_{S_D} \mathcal{L}_{adv}(S_D,S_K)
\end{equation}

\noindent{\textbf{Reconstruction Loss}}: We introduce a reconstruction loss to minimize the difference between the reconstructed feature representations of the $k$ key frames in the predicted summary video $S_K(v)$ and the input frame-level representation of those $k$ key frames in the input video $v$. Let $\Lambda_{K}$ be a set of $k$ indices indicating which $k$ frames in the input video are selected in the summary. In other words, if $f\in\Lambda_{k}$, the $f$-th frame in the input video is a key frame. We can define this reconstruction loss as:
\begin{equation}\label{eq:l_reconst}
\mathcal{L}_{reconst}(S_K(v), v) = \frac 1 {k} \sum_{t=1}^{k} \| S_K(v)^{t} - v^{f_t} \|_2^2
\end{equation}
where $S_K(v)^{t}$ and $v^{f_t}$ are the features of the $t$-th frame in the output summary video $S_K(v)$ and the $f_t$-th frame (i.e. $f_t \in \Lambda_{k}$) of the input video $v$, respectively. The intuition behind this loss is to make the reconstructed feature vectors of the key frames in the summary video $S_K(v)$ similar to the feature vectors of those frames in the input video $v$.

\noindent{\textbf{Diversity Loss}}: It is desirable in video summarization that the frames in the summary video have high visual diversity~\cite{mahasseni17_cvpr,zhang16_eccv,zhou2018_aaai}. To enforce this constraint, we apply a repelling regularizer \cite{zhao2016energy} that encourages the diversity in the output summary video $S_K(v)$ for the given input video $v$. This diversity loss is defined as: 
\begin{equation}\label{eq:l_div}
\mathcal{L}_{div}(S_K(v)) = \frac{1}{k(k-1)}\sum_{t=1}^{k}\sum_{\substack{t'=1, t'\neq t}}^{k} \frac{(S_K(v)^t)^T \cdot {S_K(v)^{t'}}}{\Vert S_K(v)^{t} \Vert _2 \Vert S_K(v)^{t'} \Vert _2}
\end{equation}
where $S_K(v)^{t}$ is the frame-level reconstructed feature representation of frame $t$ in the summary video $S_K(v)$. We aim to minimize $\mathcal{L}_{div}(S_K(v))$, so that the selected $k$ key frames are visually diverse.

\noindent{\textbf{Final Loss}}: Our final loss function is:
\begin{eqnarray}\label{eq:f_obj}\nonumber
\mathcal{L}(S_K,S_D) &=& \mathcal{L}_{adv}(S_D,S_K) + \mathcal{L}_{reconst}(S_K(v),v) \\
&& +\beta \, \mathcal{L}_{div}(S_K(v))
\end{eqnarray}
where $\beta$ is a hyperparameter that controls the relative importance of the visual diversity. The goal of the leaning is to find the optimal parameters $\Theta_{S_K}^*$ and $\Theta_{S_D}^*$ in $S_K$ and $S_D$, respectively. We can express this as the following:
\begin{equation}\label{eq:goal}
\Theta_{S_K}^*,\Theta_{S_D}^* =  \underset{\Theta_{S_K},\Theta_{S_D}}{\arg\min} \; \mathcal{L}(S_K,S_D)
\end{equation}

For brevity, we use \texttt{UnpairedVSN} to denote our unpaired video summarization network that is learned by Eq.~\ref{eq:goal}. In Fig. \ref{fig:network}(c) and Fig. \ref{fig:network}(d), we show the training scheme of $S_K$ and $S_D$ in our model \texttt{UnpairedVSN}.

\subsection{Learning with Partial Supervision} 
In some cases, we may have a small amount of paired videos during training. We use $V_p$ ($V_p\subset V$) to denote this subset of videos for which we have the ground truth summary videos. Our model can be easily extended to take advantage of this partial supervision. In this case, we apply an additional objective $\mathcal{L}_{psup}$ on the output of FCSN in the key frame selector network $S_K$. Suppose a training input video $v \in V_p$ has $T$ frames, $\delta_{t,l}$ is the score of the $t$-th frame to be the $l$-th class (key frame or non-key frame) and $l_{t}$ is the ground truth binary key frame indicator. We define $\mathcal{L}_{psup}(v)$ as:
\begin{equation}
\mathcal{L}_{psup}(v) = -\frac{1}{T}\sum_{t=1}^{T} \log\Big(\frac{\exp(\delta_{t,l_{t}})}{\sum_{l=1}^{2}\exp(\delta_{t,l})}\Big)
\end{equation}
Our learning objective in this case is defined as:
\begin{eqnarray}\label{eq:l_psup}\nonumber
\mathcal{L}(S_K,S_D)&=&\mathcal{L}_{adv} + \mathcal{L}_{reconst} + \beta \, \mathcal{L}_{div}\\
&& + \gamma \cdot \mathbbm{1}{(v)} \cdot \mathcal{L}_{psup}
\end{eqnarray}
where $\mathbbm{1}{(\cdot)}$ is an indicator function that returns $1$ if $v \in V_p$, and $0$ otherwise. This means that $\mathcal{L}_{psup}$ is considered if the video $v$ is an instance in $V_p$ for which we have the ground-truth summary video. The hyperparameters $\beta$ and $\gamma$ control the relative importance of the diversity and supervision losses, respectively. We denote this variant of our model as \texttt{UnpairedVSN}$_{psup}$.

\section{Experiments}\label{sec:exp}

\subsection{Setup}\label{sec:setup}
\begin{table*}[!ht]
	\small
	\begin{center}
		\begin{tabular}{c|c|c|c}
			\hline
			Dataset & No. of videos & Content &  Ground truth annotation type\\
			\hline
			SumMe~\cite{gygli14_eccv} & 25 & User videos &  Interval-based shots and frame-level score\\
			TVSum~\cite{Song_2015_CVPR} & 50 & YouTube videos &  Frame-level importance score\\
			YouTube~\cite{de2011vsumm}\textsuperscript{\dag} & 39 & Web videos &  Collection of key frames \\
			OVP~\cite{ovp} & 50 & Various genre videos & Collection of key frames \\
			\hline
		\end{tabular}
	\end{center}
	\caption{Key characteristics of different datasets used in our experiments. \textsuperscript{\dag}The YouTube dataset has 50 videos, but we exclude (following \cite{gong14_nips,zhang16_eccv}) the 11 cartoon videos and keep the rest.}
	\label{table:datasets}
\end{table*}

\noindent{\textbf{Data and Setting}}: We conduct evaluation on two standard video summarization datasets: SumMe \cite{gygli14_eccv} and TVSum \cite{Song_2015_CVPR}. These datasets have 25 and 50 videos, respectively. Since these datasets are very small, we use another two datasets, namely the YouTube \cite{de2011vsumm} ($39$ videos) and the OVP dataset \cite{ovp} ($50$ videos), to help the learning. Table \ref{table:datasets} shows the main characteristics of the datasets. We can observe that these datasets are diverse, especially in terms of ground truth annotations. We follow prior work \cite{gong14_nips,zhang16_eccv} to convert multiple ground truths with different format to generate a single keyframe-based annotation (a binary key frame indicator vector \cite{zhang16_eccv}) for each training video.

From Table \ref{table:datasets}, we can see that we have in total $164$ videos available for experiments. When evaluating on the SumMe dataset, we randomly select $20$\% of SumMe videos for testing. We use the remaining $80$\% of SumMe videos and all the videos in other datasets (i.e., TVSum, YouTube and OVP) for training. We create the \textit{unpaired} data from the training subset by first randomly selecting $50$\% of the raw videos (ignoring their ground truth summaries) and then selecting the ground truth summaries (while ignoring the corresponding raw videos) of the remaining $50$\% videos. In the end, we obtain a set of raw videos and a set of real summary videos, where there is no correspondence between the raw videos and the summary videos. We follow the same strategy to create the training (unpaired) and testing set when evaluating on the TVSum dataset.

\noindent{\textbf{Features}}: Firstly, we uniformly downsample every video to $2$ fps. Then we use $pool5$ layer of the pretrained GoogleNet \cite{szegedy15_cvpr} to extract $1024$-dimensional feature representation of each frame in the video. Note that our feature extraction follows prior work \cite{mahasseni17_cvpr,ours18_eccv,zhang16_eccv,zhou2018_aaai}. This allows us to perform a fair comparison with these works.

\noindent{\textbf{Training Details}}: We train our final model (\texttt{UnpairedVSN}) from scratch with a batch size of $1$. We use the Adam optimizer \cite{kingma15_iclr} with a learning rate of $0.00001$ for the key frame selector network ($S_K$). We use the SGD \cite{bottou2010_sgd} optimizer with a learning rate of $0.0002$ for the summary discriminator network ($S_D$). We set $\beta = 1$ for SumMe and $\beta = 0.001$ for TVSum in Eq.~\ref{eq:f_obj}. Additionally, we set $\beta$  and $\gamma$ to $0.001$ for SumMe and TVSum in Eq.~\ref{eq:l_psup}.


\noindent{\textbf{Evaluation Metrics}}: We evaluate our method using the keyshot-based metrics as in previous work \cite{mahasseni17_cvpr,zhang16_eccv}. Our method predicts summaries in the form of key frames. We convert these key frames to key shots (i.e., an interval-based subset of video frames \cite{gygli14_eccv,gygli15_cvpr,zhang16_eccv}) following the approach in \cite{zhang16_eccv}. The idea is to first temporally segment the videos using KTS algorithm \cite{potapov14_eccv}. If a segment contains a key frame, we mark all the frames in that segment as $1$, and $0$ otherwise. This process may result in many key shots. In order to reduce the number of key shots, we rank the segments according to the ratio between the number of key frames and the length of segment. We then apply knapsack algorithm to generate keyshot-based summaries that are at most $15$\% of the length of the test video \cite{gygli14_eccv,gygli15_cvpr,Song_2015_CVPR,zhang16_eccv}. The SumMe dataset has keyshot-based ground truth annotation, so we directly use it for evaluation. The TVSum dataset provides frame-level importance scores which we also convert to key shots as done by \cite{mahasseni17_cvpr,zhang16_eccv} for evaluation.

Given a test video $v$, let $X$ and $Y$ be the predicted key shot summary and the ground truth summary, respectively. We compute the precision ($P$), recall ($R$) and F-score ($F$) to measure the quality of the summary as follows:
\begin{eqnarray}
P=\frac{\text{overlap in $X$ and $Y$}}{\text{duration of $X$}}, R=\frac{\text{overlap in $X$ and $Y$}}{\text{duration of $Y$}}
\end{eqnarray}
\begin{equation}
F=\frac{2 \times P \times  R}{P + R}
\end{equation}
We follow the evaluation protocol of the datasets (SumMe \cite{gygli14_eccv,gygli15_cvpr} and TVSum \cite{Song_2015_CVPR}) to compute the F-score between the multiple user created summaries and the predicted summary for each video in the datasets. Following prior work  \cite{mahasseni17_cvpr}, we run our experiments five times for each method and report the average performance over the five runs.

\subsection{Baselines}\label{sec:baselines}
Since our work is the first attempt to learn video summarization using unpaired data, there is no prior work that we can directly compare with. Nevertheless, we define our own baselines as follows:

\textbf{Unsupervised SUM-FCN}: If we remove the summary discriminator network from our model, we can learn video summarization in an unsupervised way. In this case, our learning objective is simply $\mathcal{L}_{reconst} + \mathcal{L}_{div}$. This is equivalent to the unsupervised SUM-FCN in \cite{ours18_eccv}. We call this baseline model \texttt{SUM-FCN}$_{unsup}$. Note that \texttt{SUM-FCN}$_{unsup}$ is a strong baseline (as shown in \cite{ours18_eccv}) since it already outperforms many existing unsupervised methods (\cite{de2011vsumm,khosla2013large,li2010multi,mahasseni17_cvpr,Song_2015_CVPR,zhao2014quasi}) in the literature.
 
\textbf{Model with Adversarial Objective}: We define another baseline model where we have the summary discriminator network $S_D$ and the key frame selector network $S_K$, but the objective to be minimized is $\mathcal{L}_{reconst} + \mathcal{L}_{adv}$ (i.e., we ignore $\mathcal{L}_{div}$). We refer to this baseline model as \texttt{UnpairedVSN}$_{adv}$.

\subsection{Main Results}\label{sec:main_results}


\begin{table}[!ht]
	\scriptsize
	\begin{center}
		\begin{tabular}{c|c|c|c}
			\hline
			& SUM-FCN$_{unsup}$ \cite{ours18_eccv} & \texttt{UnpairedVSN}$_{adv}$ & \texttt{UnpairedVSN}\\
			\hline
			F-score & 44.8 & 46.5 & \textbf{47.5}\\
			Precision& 43.9 & 45.0 & \textbf{46.3}\\
			Recall& 46.2 & 49.1 & \textbf{49.4}\\
			\hline
		\end{tabular}
	\end{center}
	\caption{Performance (\%) of different methods on the SumMe dataset~\cite{gygli14_eccv}. We report summarization results in terms of three standard metrics including F-score, Precision and Recall. }
	\label{table:results_summe}
\end{table}
\begin{table}[!ht]
	\scriptsize
	\begin{center}
		\begin{tabular}{c|c|c|c}
			\hline
			& SUM-FCN$_{unsup}$ \cite{ours18_eccv} & \texttt{UnpairedVSN}$_{adv}$ & \texttt{UnpairedVSN}\\
			\hline
			F-score & 53.6 & 55.3 & \textbf{55.6}\\
			Precision & 59.1 & 61.0 & \textbf{61.1}\\
			Recall & 49.1 & 50.6 & \textbf{50.9}\\
			\hline
		\end{tabular}
	\end{center}
	\caption{Performance (\%) of different methods on TVSum~\cite{Song_2015_CVPR}.}
	\label{table:results_tvsum}
\end{table}

In Table \ref{table:results_summe}, we provide the results (in terms of F-score, precision and recall) of our final model \texttt{UnpairedVSN} and the baseline models on the SumMe dataset. Our method outperforms the baseline methods on all evaluation metrics. It is also worth noting that when our summary generator and discriminator networks are trained using unpaired data with the adversarial loss (i.e., \texttt{UnpairedVSN$_{adv}$}), we observe a significant boost in performance (1.7\%, 1.1\% and 2.9\% in terms of F-score, precision and recall, respectively) over the unsupervised baseline \texttt{SUM-FCN}$_{unsup}$. Adding an additional regularizer $\mathcal{L}_{div}$ (i.e., \texttt{UnpairedVSN}) further improves the summarization performance.

Table \ref{table:results_tvsum} shows the performance of different methods on the TVSum dataset. Again, our final method outperforms the baseline methods. Moreover, the trend in performance boost is similar to what we observe on the SumMe dataset.

Results in Table \ref{table:results_summe} and Table \ref{table:results_tvsum} demonstrate that learning from unpaired data is advantageous as it can significantly improve video summarization models over purely unsupervised approaches.

\subsection{Comparison with Supervised Methods}\label{sec:comparison}
 We also compare the performance of our method with state-of-the-art supervised methods for video summarization. Recent supervised methods \cite{mahasseni17_cvpr,ours18_eccv,zhang16_cvpr,zhang16_eccv,zhang2018_eccv,zhao2017_acmm,zhou2018_aaai} also use additional datasets (i.e., YouTube and OVP) to increase the number of paired training examples while training on the SumMe or the TVSum dataset. For example, when experimenting on SumMe, they use $20\%$ for testing and use the remaining $80\%$ videos of SumMe along with the videos in TVSum, OVP and YouTube for training. However, the main difference is that we further divide the combined training dataset to create unpaired examples (see Sec. \ref{sec:setup}). In other words, given a pair of videos (a raw video and its summary video), we either keep the raw video or the summary video in our training set. In contrast, both videos are part of the training set in supervised methods. As a result, supervised methods use twice as many videos during training. In addition, supervised methods have access to the correspondence between the raw video and the ground truth summary video. Therefore, it is important to note that the supervised methods utilize far more supervision than our proposed method. We show the comparison in Table~\ref{table:comp_supervised}. 
 
 Surprisingly, on the SumMe dataset, our final method outperforms most of the supervised methods (except \cite{ours18_eccv}) by a big margin (nearly 3\%). On the TVSum dataset, we achieve slightly lower performance. Our intuition is that if we have more unpaired data for training, we can reduce the performance gap on TVSum. To sum up, this comparison study demonstrates that our unpaired learning formulation has potential to compete with supervised approaches.

\begin{table}[h]
	\begin{center}
		\begin{tabular}{c|c|c}
			\hline
			Method &SumMe & TVSum \\
			\hline
			Zhang \etal \cite{zhang16_cvpr} & 41.3 & --\\
			Zhang \etal \cite{zhang16_eccv} (vsLSTM) & 41.6 & 57.9\\
			Zhang \etal \cite{zhang16_eccv} (dppLSTM) & 42.9 & 59.6\\
			Mahasseni \etal \cite{mahasseni17_cvpr} (supervised) & 43.6 & 61.2\\
			Zhao \etal \cite{zhao2017_acmm}\textsuperscript{\ddag} & 43.6 & 61.5\\
			Zhou \etal \cite{zhou2018_aaai} (supervised) & 43.9 & 59.8\\
			Zhang \etal \cite{zhang2018_eccv} & 44.1 & 63.9 \\
			Rochan \etal \cite{ours18_eccv} & 51.1 & 59.2 \\ \hline
			\texttt{UnpairedVSN} (Ours) & 47.5 & 55.6\\
			\hline
		\end{tabular}
	\end{center}
	\caption{Quantitative comparison (in terms of F-score \%) between our methods and state-of-the-art supervised methods on SumMe~\cite{gygli14_eccv} and TVSum~\cite{Song_2015_CVPR}. \textsuperscript{\ddag}Results are taken from \cite{zhang2018_eccv}.}
	\label{table:comp_supervised}
\end{table}

\subsection{Effect of Partial Supervision}\label{sec:p_supervision}
We also examine the performance of our model when direct supervision (i.e., correspondence between videos in $V$ and $S$) is available for a small number of videos in the training set. Our aim is to study the effect of adding partial supervision to the framework.

In this case, for the first $10$\% of original/raw videos that are fed to the key frame selector network, we use their ground truth key frame annotations as an additional learning signal (see Eq.~\ref{eq:l_psup}). Intuitively, we should be able to obtain better performance than learning only with unpaired data, since we have some extra supervision during training.

Table \ref{table:results_partial_supervision} shows the performance of our model trained with this additional partial supervision. We observe a trend of improvement (across all evaluation metrics) on both the datasets. This shows that our proposed model can be further improved if we have access to some paired data in addition to unpaired data during the training.

\begin{table}[h]
	\begin{center}
		\begin{tabular}{c|c|c}
			\hline
			  & SumMe & TVSum\\
			\hline
			F-score & \textbf{48.0} (47.5) &  \textbf{56.1} (55.6)\\
			Precision & \textbf{46.7} (46.3) & \textbf{61.7} (61.1)\\
			Recall& \textbf{49.9} (49.4) & \textbf{51.4} (50.9) \\
			\hline
		\end{tabular}
	\end{center}
	\caption{Performance (\%) of \texttt{UnpairedVSN}$_{psup}$ on the SumMe~\cite{gygli14_eccv} and TVSum~\cite{Song_2015_CVPR} datasets. In the bracket, we include the performance of our final model \texttt{UnpairedVSN} reported in Table \ref{table:results_summe} and Table \ref{table:results_tvsum} to help with the comparison.}
	\label{table:results_partial_supervision}
\end{table}

\subsection{Transfer Data Setting}\label{sec:video-wise}
In our standard data setting (see Sec. \ref{sec:setup}), it is possible that some of the unpaired examples consist of raw videos or video summaries from the dataset under consideration. In order to avoid this, we conduct additional experiments under a more challenging data setting where the unpaired examples originate totally from different datasets. For instance, if we evaluate on SumMe, we use the videos and user summaries of TVSum, OVP and YouTube to create unpaired training data, and then use the entire SumMe for testing. We follow the similar process while evaluating on TVSum. This kind of data setting is referred as transfer data setting \cite{zhang16_cvpr,zhang16_eccv}, though it has been defined in the context of fully supervised learning. We believe that this data setting is closer to real scenarios, where we may need to summarize videos from domains that are different from those used in training.

Table \ref{table:results_summe_transfer} and Table \ref{table:results_tvsum_transfer} show the performance of different approaches on SumMe and TVSum, respectively. Although we notice slight degradation in performance compared with the standard data setting, the trend in results is consistent with our findings in Sec. \ref{sec:main_results}.

\begin{table}[!ht]
	\scriptsize
	\begin{center}
		\begin{tabular}{c|c|c|c}
			\hline
			& SUM-FCN$_{unsup}$ \cite{ours18_eccv} & \texttt{UnpairedVSN}$_{adv}$ & \texttt{UnpairedVSN}\\
			\hline
			F-score & 39.5 & 41.4 & \textbf{41.6}\\
			Precision& 38.3 & 40.4 & \textbf{40.5}\\
			Recall& 41.2 & 43.6 & \textbf{43.7}\\
			\hline
		\end{tabular}
	\end{center}
	\caption{Performance (\%) of different methods on SumMe~\cite{gygli14_eccv} under transfer data setting.}
	\label{table:results_summe_transfer}
\end{table}
\begin{table}[!ht]
	\scriptsize
	\begin{center}
		\begin{tabular}{c|c|c|c}
			\hline
			& SUM-FCN$_{unsup}$ \cite{ours18_eccv} & \texttt{UnpairedVSN}$_{adv}$ & \texttt{UnpairedVSN}\\
			\hline
			F-score & 52.9 & 55.0 & \textbf{55.7}\\
			Precision & 58.2 & 60.6 & \textbf{61.2}\\
			Recall & 48.5 & 50.4 & \textbf{51.1}\\
			\hline
		\end{tabular}
	\end{center}
	\caption{Performance (\%) of different methods on TVSum~\cite{Song_2015_CVPR} under transfer data setting.}
	\label{table:results_tvsum_transfer}
\end{table}

\subsection{Qualitative Analysis}\label{sec:qual}
\begin{figure*}[ht]
	\center
	\includegraphics[width=0.94\textwidth]{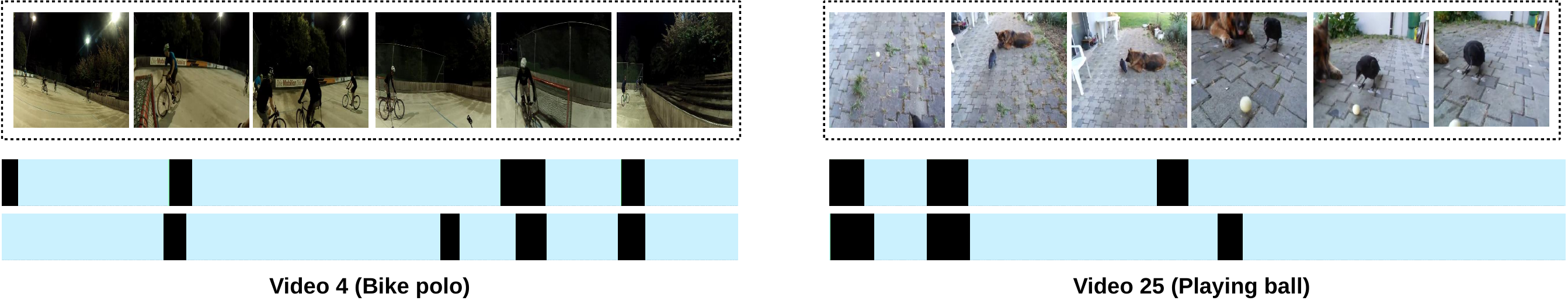}
	\caption{Two example results from the SumMe dataset \cite{gygli14_eccv}. The two bars at the bottom show the summaries produced by \texttt{UnpairedVSN} and humans, respectively. The black bars denote the selected sequences of frames, and the blue bar in background indicate the video length.}
	\label{fig:qual2}
\end{figure*}

\begin{figure*}[ht]
	\center
	\includegraphics[width=0.94\textwidth]{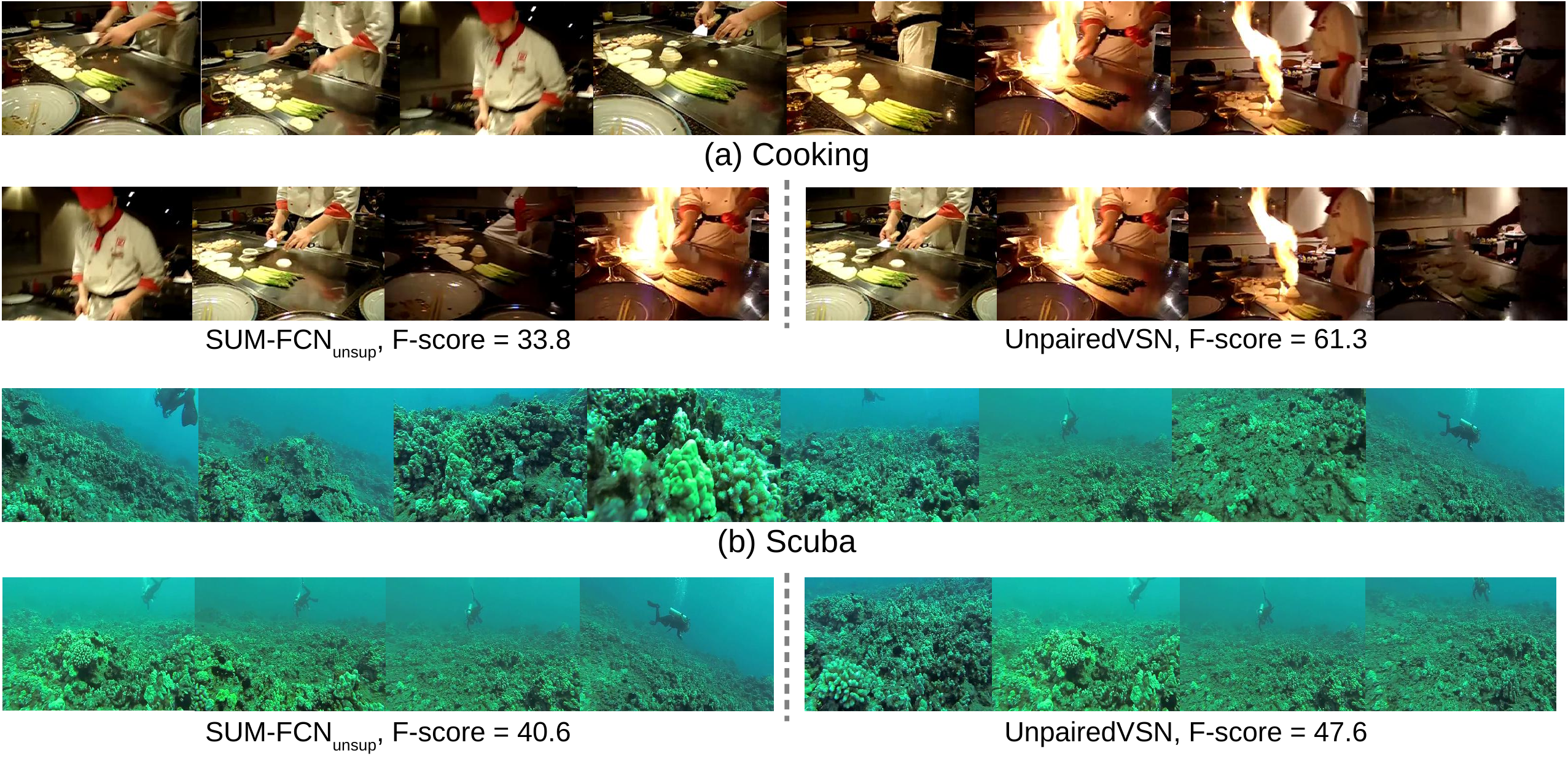}
	\caption{Example videos from SumMe \cite{gygli14_eccv} and predicted summaries by \texttt{SUM-FCN}$_{unsup}$ \cite{ours18_eccv} and \texttt{UnpairedVSN}. Frames in the first row are sampled from the video, whereas frames in the second row are sampled from the summaries generated by different approaches.}
	\label{fig:qual}
\end{figure*}

Figure \ref{fig:qual2} presents example summaries generated by our method \texttt{UnpairedVSN}. We observe that the output summaries from our approach have a higher overlap with the human generated summaries. This implies that our method is able to preserve information essential for generating optimal and meaningful summaries.

We compare the results of different approaches in Fig. \ref{fig:qual}. The first video in Fig. \ref{fig:qual}(a) is related to cooking. \texttt{SUM-FCN}$_{unsup}$ extracts the shots from the middle of the video and misses the important video shots towards the end. In contrast, we observe that \texttt{UnpairedVSN} preserves the temporal story of the video by extracting video shots from different sections while focusing on key scenes. This has resulted in better agreement with the human created summaries. The second video in Fig. \ref{fig:qual}(b) is about scuba diving. Unlike the first video, there is not a huge performance gap between \texttt{SUM-FCN}$_{unsup}$ and \texttt{UnpairedVSN}. However, it still noticeable that \texttt{SUM-FCN}$_{unsup}$ captures less diverse scenes compared with \texttt{UnpairedVSN}.


\section{Conclusion}\label{sec:conclude}
We have presented a new formulation for video summarization where the goal is to learn video summarization using unpaired training examples. We have introduced a deep learning framework that operates on unpaired data and achieves much better performance than the baselines. Our proposed method obtains results that are even comparable to state-of-the-art supervised methods. If a small number of paired videos are available during training, our proposed framework can be easily extended to take advantage of this extra supervision to further boost the performance. Since unpaired training data are much easier to collect, our work offers a promising direction for future research in video summarization. As future work, we plan to experiment with large-scale unpaired videos collected in the wild.

\noindent{\bf Acknowledgments}: The authors acknowledge financial support from NSERC and UMGF funding. We also thank NVIDIA for donating some of the GPUs used in this work.

{\small
\bibliographystyle{ieee_fullname}
\bibliography{wang}
}

\end{document}